**Original Investigation**

**Title: Improving Surgical Risk Prediction Through Integrating Automated Body Composition Analysis: a Retrospective Trial on Colectomy Surgery**

Hanxue Gu, BS[1]; Yaqian Chen, MS[1]; Jisoo Lee, MS[2]; Diego Schaps, MD, MPH[3]; Regina Woody, BSN[3]; Roy Colglazier, MD[4]; Maciej A. Mazurowski, PHD[1,2,4,5]; Christopher Mantyh, MD[3]

[1]Department of Electrical and Computer Engineering, Duke, NC, 27703, USA; [2]Department of Biostatistics and Bioinformatics, Duke University; [3]Department of Surgery Duke University School of Medicine; [4]Department of Radiology, Duke University; [5]Department of Computer Science, Duke University

Corresponding author: Hanxue Gu, [1]Department of Electrical and Computer Engineering, Duke, NC, 27703, USA, hanxue.gu@duke.edu

Word count: 2998

*Abstract:*

**Importance:** Body composition measurements may help predict surgical outcomes, but current approaches are not optimized across body regions or outcome types. Their added value beyond established clinical factors remains unclear, highlighting a critical gap.

**Objective:** To evaluate whether preoperative body composition metrics automatically extracted from CT scans can predict postoperative outcomes after colectomy, either alone or combined with clinical variables or existing risk predictors.

**Design:** Retrospective cohort study.

**Setting:** Tertiary health system with three hospitals.

**Participants:** 3,560 patients who underwent colectomy between January 1, 2010 and December 31, 2023.

**Exposure:** Risk prediction models using body composition metrics alone, the American College of Surgeons National Surgical Quality Improvement Program (NSQIP) Risk Calculator alone, or a combination of both.

**Main outcomes and measures:** The primary outcome was the predictive performance for 1-year all-cause mortality following colectomy. A Cox proportional hazards model with 1-year follow-up was used, and performance was evaluated using the concordance index (C-index) and Integrated Brier Score (IBS).

Secondary outcomes included postoperative complications, unplanned readmission, blood transfusion, and severe infection, assessed using AUC and Brier Score from logistic regression. Odds ratios (OR) described associations between individual CT-derived body composition metrics and outcomes. Over 300 features were extracted from preoperative CTs across multiple vertebral levels, including skeletal muscle area, density, fat areas, and inter-tissue metrics. NSQIP scores were available for all surgeries after 2012.

**Results:** A total of 1,623 patients who underwent colectomy were included. Of these, 476 patients (years 2010–2015) formed the development subset, while 1,147 patients (years 2016–2023) were used for validation. Optimal anatomic locations for body composition measurement varied by the predicted outcome. Sex-normalized skeletal muscle density at the L3 vertebral level was the strongest independent predictor of 1-year mortality (OR: 0.42, AUC: 0.71), while it measured at the T12 level better predicted postoperative complications (OR: 0.62, AUC: 0.63). Integrating image-based scores with clinical variables (BMI, Age, etc.) improved predictive accuracy compared to clinical variables alone: mortality (C-index 0.80 vs. 0.73, respectively) and other complications (AUC: 0.70 vs. 0.65), respectively. Adding sex-normalized skeletal muscle density to the NSQIP Surgical Risk Calculator provided a modest but significant improvement (C-index: 0.87 vs. 0.86, p<0.01). The larger benefit was observed among patients deemed low-risk by the NSQIP risk calculator (predicted mortality risk <5%), in whom the C-index increased by 4% (p<0.001), suggesting that incorporating image-based metrics can help identify at-risk individuals who might otherwise be overlooked by existing risk assessment tools.

**Conclusions and relevance:** Body composition measurements strongly predict colectomy outcomes, especially in patients thought to be low-risk preoperatively.

**Introduction**

Colectomy is a common surgical procedure performed for a range of indications, including colorectal cancer, diverticulitis, inflammatory bowel disease, and pre-malignant lesions[1,2]. However, it is associated with significant morbidity and even mortality[1]. Nearly one-third of patients experience postoperative complications, including infection, anastomotic leak, or organ dysfunction as well as other life-threatening consequences[3–5]. This emphasizes the critical need for effective preoperative risk stratification for guiding surgical decision-making, tailoring of preoperative optimization strategies, and planning of postoperative care[6]. Despite this need, existing surgical risk prediction tools, including the widely used American College of Surgeons (ACS) National Surgical Quality Improvement Program (NSQIP) Surgical Risk Calculator, present significant limitations[7]. Specifically, ACS-NSQIP has been criticized for underestimating patient-specific risks and failing to capture the physiologic heterogeneity of surgical candidates. Frailty, marked by diminished physiologic reserve, emerges as a strong predictor of adverse surgical outcomes[8,9]. Traditional frailty indices, such as the Fried Frailty Index[10,11], rely heavily on subjective assessments, which may introduce bias and lack standardized, objective measurements. On the other hand, general frailty indicators, such as Body Mass Index (BMI) and age, often fail to reflect the variance in muscle and fat distribution that may influence surgical resilience. Emerging evidence highlights that cross-sectional imaging offers a promising opportunity to fill this gap[12–18]. Computed tomography (CT)-derived body composition metrics such as skeletal muscle density and visceral fat area provide objective, and quantifiable insights into a patient's physiologic reserve[19,20]. However, the optimal anatomic regions for measurement, most informative metrics, and their additive predictive value over existing clinical models remain uncertain, warranting further investigation.

To address this, we developed and validated risk prediction models utilizing automated CT-based body composition analysis in patients undergoing colectomy – including total colectomy, formal anatomical resections, and segmental resection. By systematically comparing these image-based features with conventional clinical predictors, we aim to identify the most prognostic imaging metrics for different surgical outcomes and to assess their independent value in improving preoperative risk stratification beyond current models and frailty scores, seeking to identify a novel imaging biomarker automatically extracted from CT-images that may enhance surgical decision optimization.

## Methods

### Retrospective cohort of colectomy surgical patients

We conducted a retrospective cohort study of patients who underwent colectomy at three attending hospitals (Hospital A: Duke University Hospital; Hospital B: Duke Raleigh Hospital, and Hospital C: Duke Regional Hospital) from 2010 to 2023, using the ACS NSQIP database. inclusion criteria were: (1) patients who underwent formal anatomic resection (e.g., left or right hemicolectomy), segmental colectomy, or total colectomy whether via an open or minimally invasive technique; (2) a minimum of 30 days of postoperative follow-up to assess short-term outcomes; and (3) availability of axial chest, abdomen, or pelvis CT scans performed within 90 days preoperatively and showing detectable L3 vertebra. For patients with multiple operations, only the first procedure was selected. To ensure accurate body composition analysis, the smallest slice thickness and 'original' image types were selected over reconstructed scans. The full data selection process is illustrated in [eFigure 1] in Supplement 1. Demographic variables including age, sex, and BMI were obtained from the electronic medical record (EMR).

Operations performed at Hospital A between 2010 and 2015, were used to develop, analyze, and select body composition metrics and predictive models. Operations from 2016 to 2023 were reserved for model validation, incorporating data from two additional hospitals in the health (Hospital B and Hospital C) for external validation.

### Automatic image-based frailty scores extraction

A comprehensive set of body composition metrics was extracted in both 2D and 3D from the T12 to L4 vertebral region, a standard coverage area in abdominal CT scans, by a deep learning-based automatic segmentation method (*eMethod 1* in Supplement 1)[21]. 2D scores quantified tissues at specific vertebral levels, while 3D scores measured volumetric distributions across T12 to L4. These scores were categorized into three groups. First, **direct scores** included absolute tissue measurements, such as Skeletal Muscle Area (SMA), Subcutaneous Fat Area (SFA), Visceral Fat Area (VFA), inter-/intra-Muscular Fat Area (MFA), and body area (BODY, all non-background pixels with Hounsfield Unit (HU)>-1000). In 2D analysis, pixel counts within regions were converted to area (mm²), while 3D analysis computed volume (mm³) from T12 to L4. Additionally, Skeletal Muscle Density (SMD) was calculated as mean HU within the muscle regions. Second, **within-body-derived scores** assessed relationships between muscle and fat compartments. The combined Skeletal Muscle and inter-/intra-Muscular Fat Area (SMFA) and its density (SMFD) represented total muscle and fat areas along with their average HU values. Fat distribution was further characterized by Muscle-to-Fat Ratio (MFR), VFA/SFA, SFA/SMA, VFA/SMA, and MFA/SMA. Additionally, fat-to-body area ratios (SFA/BODY, VFA/BODY, and MFA/BODY) normalized fat compartments to body area. Lastly, **demographic-adjusted scores** accounted for distribution differences in sex or other patient features. The Skeletal Muscle Index (SMI) and Fat Mass Index (FMI) were computed by normalizing SMA and SFA at L3 (cm²) by height squared (m²). The Sarcopenic Obesity Index (SOI) was computed as SMI divided by VFA at L3. Since SMI, FMI, and SOI are clinically defined at L3, these were restricted to this level, while all other 2D metrics were measured at multiple vertebral levels[22–24]. To account for sex differences, sex-normalized scores (N_SMA, N_SMD, etc.) were calculated using Z-score normalization, considering sex-specific thresholds for sarcopenia and myosteatosis[25]. Details and illustrations of these scores are shown in [Figure 1, part (a)], and in *eTable 1* in Supplementary 1.

**Surgical variables collection**

Preoperative, perioperative, and postoperative variables were collected. Preoperative data included patient demographics (sex, age, race) and baseline risk factors such as BMI, functional status, and comorbidities. Perioperative data included type of operation performed and emergency status. Postoperative data focused on 30-day outcomes, including *mortality*, *any complication*, *serious complication*, and *unplanned readmission*, following the NSQIP definitions[26], as well as other outcomes, including sepsis; septic shock; Clostridioides difficile infection; pulmonary complications (unplanned intubation, prolonged mechanical ventilation >48 hours, pneumonia); cardiac complications (myocardial infarction and cardiac arrest requiring CPR); renal complications (renal insufficiency and hemodialysis); severe infections (deep incisional and organ/space surgical site infections); neurological events (stroke); thromboembolic events (venous thrombosis and pulmonary embolism); unplanned return to OR; and postoperative transfusion (*eTable 2*). Short-term outcomes were recorded as part of the NSQIP standardized 30-day postoperative follow-up protocol. One-year mortality was extracted from the medical record based on most recent documented follow-up. More details are available in *eMethod 2* in Supplement 1.

**Existing risk model's risk assessment collection**

ACS NSQIP Surgical Risk assessments were collected for patients undergoing colectomy since 2012 when standardized NSQIP input variables (21 variables) start collecting at the three hospitals using the NSQIP ALLCLASS model, focusing on mortality outcomes.

**Outcomes and Study Endpoint**

The primary endpoint was all-cause mortality within 1 year postoperatively following colectomy, with 30-day and 1-year mortality serving as key time points for analysis. Other secondary endpoints included 30-day any complication, major complication, unplanned readmission, and several specific postoperative complications selected based on their incidence in the development cohort.

**Statistical Analysis**

**Model development**

**Image-based frailty score selection:** The selection of body composition scores followed two primary objectives: (1) identifying the most predictive vertebral level for each metric, and (2) determining the most predictive subset of metrics overall. Univariate logistic regression was performed on the development cohort to evaluate the association between each body composition score and each outcome. For metrics available at multiple vertebral levels, the level with the highest Area Under the Receiver Operating Characteristic curve (AUC) was selected. To minimize redundancy and multicollinearity, only the most predictive and independent metrics were retained. Specifically, metrics were ranked by AUC, and among highly correlated metrics (Corr>0.8), only the top-performing score was kept. For each selected metric, we reported the odds ratio (OR), 95% confidence interval (CI), AUC, and p-value.

To evaluate potential confounding by commonly available frailty-related variables (age group, BMI category, smoking status, American Society of Anesthesiologists [ASA] physical and functional class), each image-based score was assessed in a multivariable logistic regression model adjusting for these covariates. Image-based scores that remained significant (p<0.1) after adjustment were retained; others were excluded from further modeling. The complete image-based score selection pipeline is shown in [Figure 1, part (b)].

**Multivariable prediction model development:** To assess mortality, two endpoints were evaluated: *1-year* and *30-day* all-cause mortality. A Cox proportional hazards model was built using 1-year follow-up data to model time-to-event outcomes. For binary surgical outcomes, a multivariable regression model was constructed.

For each outcome, we developed three core models: **IMG-only**, which included only image-based body composition scores; **CLIN-only**, which included easily accessible clinical confounders such as age group, BMI category, smoking status, functional status, and ASA class; and **IMG+CLIN**, which combined both image-based scores and clinical confounders. Additionally, for 1-year mortality, we developed two survival models incorporating NSQIP-predicted risk: **IMG+NSQIP**, which combined image-based scores with NSQIP risk prediction, and **NSQIP-only**, which included only NSQIP risk prediction. Predictor variables for each model were selected using backward stepwise elimination, removing variables with p > 0.1. Model development and variable selection were conducted exclusively on the development dataset.

Model validation

**Multivariable Prediction of Image-scores alone or with cofounders:** We validated the performance of multivariable models **IMG-only, CLIN-only** and **IMG+CLIN** for all outcomes using the hold-out test set. Predictive performance of each model was quantified using the concordance index (C-index) for discrimination and Integrated Brier Score (IBS) for calibration in Cox proportional hazards models for *mortality*. We used the area under the receiver operating characteristic curve (AUC) for discrimination and Brier score (BS) for calibration in logistic regression models for *complication outcomes*. Statistical testing was performed using bootstrap resampling (1,000 iterations) to compare model performance. To further characterize model behavior, we analyzed feature contributions: for survival models, we assessed the hazard ratios (HRs) for each predictor; and for binary outcome prediction models, we visualized feature importance using SHAP (SHapley Additive exPlanations) values.

**Comparison with NSQIP risk scores:** To assess the added prognostic value of image-based frailty scores beyond existing clinical risk estimates, we performed subgroup analyses within the independent validation set. For mortality outcomes, two Cox models were built: one incorporating both NSQIP and image-based scores (NSQIP+IMG) and one using NSQIP scores alone (NSQIP-only). Model performance was assessed on the validation set using C-index and IBS. Kaplan-Meier survival analysis was performed to compare time-to-event differences across stratified risk groups. Patients were categorized into low- or high-strata image-based score, as well as by low- or high-risk NSQIP risk groups score using the median score as the cutoff.

# Results

**Patient Population and characteristics**

**Development cohort:** Of 976 patients who underwent colectomy between 2010–2015, 474 (48.6%) met inclusion criteria (mean age: 62±14.2 years). The majority underwent laparoscopic approaches (60.8%) and 16.2% of cases were emergency cases. Thirty-day and 1-year mortality rate were 4.0% and 9.5%, respectively. Postoperative transfusion (16.5%) was the most commonly observed complication, followed by *unexpected readmission* (13.9%), *severe infections* (12.7%) and *pulmonary complication* (6.5%). These complications were selected for further modeling due to their relatively high frequency.

**Validation cohort:** Of 2,584 patients who underwent colectomy between 2016 and 2023, a validation subset of 1,147 patients was retained based on inclusion criteria. Although two additional hospitals contributed patients to the validation cohort, the distribution of demographic characteristics, baseline risk factors and surgical outcomes remained relatively consistent with those in the development cohort (*eTable 3* in Supplement 1).

**Model development**

**Image-based frailty score selection**

**Primary Endpoint – Mortality:** Univariable logistic regression was performed to assess the predictive performance (AUC) of each body composition metric for 1-year and 30-day mortality across different vertebral levels within the development cohort [Figure 2, part (a)]. Optimal vertebral level varied by body composition scores. For *1-year mortality*, sex-normalized skeletal muscle density (N_SMD) performed best at L3, aligning with prior findings[15], while muscle-to-fat ratio (MFA/SMA) had its highest predictive value at L1. For *30-day mortality*, AUC for N_SMD measured in 3D volume (from T12 to L4) was slightly higher than its 2D version at L3 and was 0.78 vs. 0.76, respectively ($p>0.1$). However, due to the limited number of 30-day mortality cases, we still selected L3 as the optimal level for subsequent mortality prediction due to its greater stability across endpoints.

After identifying the optimal vertebral level for each metric, we removed redundant features by excluding those with high collinearity (Pearson $r > 0.8$). The final univariable prediction performance of these selected metrics is summarized in *Table 1*. Among all image-based scores, N_SMD was the strongest predictor for 1-year mortality (AUC of 0.71, OR: 0.42 [0.30, 0.58], $p<0.001$), indicating that lower N_SMD values were significantly associated with increased mortality risk. Other image-based scores exhibited a noticeable performance gap, with the next highest AUC values around 0.6. The predictive effect of N_SMD was consistent across both mortality time points, with odds ratios of 0.42 for 1-year and 0.40 for 30-day mortality. In contrast, the normalized visceral-to-subcutaneous fat area ratio (N_VFA/SFA, L1) had a stronger association with 30-day mortality (OR: 0.54, 95%CI [0.29,1.0]) than with 1-year mortality (OR: 0.72, 95%CI [0.5,1.04]).

To assess the independence of these metrics from other clinical risk factors, we adjusted for age group, BMI category, smoking status, functional status, and ASA class using multivariable logistic regression. Both

N_SMD (L3) and N_VFA/SFA (L1) remained significant predictors of *mortality* after adjustment (adjusted p=0.000) and though they were significantly associated with some confounders (*eTable 4* in Supplement 1), they still demonstrated independent predictive power. Other image-based metrics like sex-normalized skeletal muscle area (N_SMA) and sex-normalized visceral fat to body ratio (N_VFA/BODY) were predictive in univariate models but did not retain statistical significance after adjustment (adjusted p>0.1). This suggests that their associations with mortality can be explained by other clinical variables.

**Secondary Endpoints:** The univariable prediction results of secondary endpoints at each vertebral level are shown in [Figure 2, part (b)]. After feature selection, sex-normalized skeletal muscle and fat averaged density (N_SMFD) was the most predictive image-based frailty metric, with its optimal measurement at T12, achieving an AUC of 0.63 (p<0.001) for *any complication* and 0.61 (p<0.001) for *serious complication* (Table 1). *Pulmonary complication* was best predicted at N_SMD (L3_L4) (AUC=0.78, p<0.0001), and *unplanned readmission* was best predicted by N_MFA/BODY at L1-L2 (AUC=0.61, p=0.01). For *postoperative transfusion*, N_SMD at L2-L3 remained the most effective predictor, whereas *severe infection* was best predicted by SMA at L3-L4 (AUC=0.60, p=0.01).

Multivariable logistic regression was also performed for each outcome, adjusting for the same set of confounders (age group, BMI category, smoking status, functional status and ASA class). N_SMD (T12) and VFA (L4) remained significant predictors of *any complication* and *major complications* after adjustment (adjusted p<0.1). While VFA/SFA measured at L4 was initially predictive in univariate analysis, it lost significance after adjustment, (p>0.2), suggesting its predictive value was mostly driven by clinical variables, e.g. smoking status and age. For other outcomes, the image-based scores that remained significant (adjusted p<0.1) were: (1) *Pulmonary complication*: N_SMD (L3-L4); (2) *Unplanned readmission,* No image-based scores remained; (3) For *postoperative transfusion*: N_SMD (3D), SFA/Body (L4), and N_MFA (L1-L2); and (4) For *severe infection*: No image-based scores remained.

**Model validation**

**Models with image-based scores alone or with cofounders:** The multivariable Cox-Hazard model using **only** image-based features (**IMG-only**) achieved a C-index of 0.70 and an integrated Brier Score (IBS) of 0.07. Logistic regression models achieve AUCs and Brier Scores of 0.58 and 0.24 for *any complication*, 0.56 and 0.25 for *serious complication*, 0.66 and 0.26 for *postoperative transfusion,* respectively [Figure 2]. Adding image-based scores to clinical variables (**IMG+CLIN**) improved prediction performance, with a C-index of 0.80 for *mortality* and AUCs of 0.64, 0.61, 0.78, and 0.70 for the respective secondary outcomes. Statistical testing using bootstrap sampling confirmed that integrating image-based scores with clinical variables significantly improved prediction performance compared with models including clinical variables alone (CLIN-only) ($p<0.05$).

**Comparison with NSQIP risk scores:** For 1-year follow-up mortality, the multivariable Cox proportional hazards model applied to the validation set achieved a modest but significant improvement (0.87 vs. 0.86, $p<0.0001$) of C-index comparing **IMG+NSQIP** and **NSQIP-only**. This improvement is more dominant in the low-NSQIP risk group (0.82 vs. 0.78, $p<0.001$), indicating a better risk prediction specifically among patients initially classified as low risk by NSQIP [Figure 4, part (b)]. Kaplan-Meier survival analysis of mortality demonstrates that patients with higher NSQIP risk consistently exhibit lower survival probabilities across all subgroups, and this pattern holds for both short-term and long-term mortality [Figure 4, part (c)]. Among patients classified as low risk by NSQIP, overall survival exceeded 95%. However, in the subset of these patients who experienced unexpected mortality, low muscle density was consistently observed ($p < 0.01$), suggesting that muscle density may help identify high-risk individuals who are otherwise underestimated by clinical risk models.

Discussion

**Primary Findings and Interpretation:** To our knowledge, this is the first study to systematically evaluate and integrate automated body composition analysis into surgical risk prediction models for patients undergoing colectomy. By extracting image-derived metrics from preoperative CT scans using deep learning, we assessed their added value when combined with both general clinical variables and the ACS NSQIP surgical risk calculator. Our results show that combining selected body composition features with standard clinical variables (e.g., BMI, age, ASA class) improved mortality prediction, with a combined model achieving a C-index of 0.80 on the separate validation set — significantly outperforming clinical variables alone. When image-derived scores were added to the ACS NSQIP calculator, the combined model achieved a C-index of 0.87, representing the best overall performance among all tested models. These findings suggest body composition metrics extracted from cross-sectional images offer meaningful and non-redundant information about patients' physiological vulnerability for surgical risk stratification.

Importantly, the greatest additive prediction power and clinical utility of integrating image-based metrics into risk prediction models was observed among patients labeled as low-risk by the NSQIP calculator. Within this subgroup, the inclusion of muscle density identified higher-than-anticipated risk individuals who would have been missed by traditional predictors. Though these patients are a relatively small proportion of the population experiencing mortality, additional detection of these often-overlooked patients is still meaningful as it can inform preoperative management and informed consent discussions. Our model allows for a more nuanced approach to counseling low-risk patients pre-operatively, as the consenting process for this group has been largely generic and non-tailored with comparison to high-risk patients. Our feature selection process identified sex-normalized skeletal muscle density (N_SMD) at L3 vertebral level as the most robust imaging predictor across mortality endpoints, with an AUC of 0.71 and an odds ratio of 0.42. These results are consistent with prior literature highlighting the association between low muscle quality and adverse surgical outcomes[15,27]. However, unlike previous studies only paying attention to a narrow set of image-based scores alone, we systematically compared (1) a broader set of body composition metrics measured at different vertebral levels; and (2) a multivariable adjustment to evaluate their independent contribution beyond existing clinical variables. Notably, many image-based metrics traditionally evaluated, such as skeletal muscle index (SMI) and visceral-to-subcutaneous fat ratio, were found to be predictive in univariate analysis but lost significance after accounting for confounding with basic clinical variables[19,28,29]. These findings suggest that some body composition metrics overlap significantly with existing clinical variables, but there are still imaging-derived features that provide independent predictive value.

**Body Composition and Complication Outcomes:** Beyond mortality, we also explored the relationship between body composition features and postoperative complications. Distinct image metrics were associated with specific complication subtypes. For instance, lower visceral fat area (VFA) at L4 was associated with an increased risk of major complications, while higher subcutaneous fat area to body size ratio (SFA/BODY) at L1 level predicted a greater likelihood of transfusion [Figure 3]. Notably, sex-normalized skeletal muscle density (N_SMD) remained a consistent predictor across various complications. These findings suggest that different body composition scores at various anatomic levels may reflect distinct physiologic function, that would predispose patients to particular complication risk. For simplicity, we recommend incorporating N_SMD and VFA as key image-derived metrics in pre-operative assessment, alongside clinical variables such as ASA classification, BMI and functional status. These combined inputs can provide an improved general complication risk stratification.

**Clinical Integration and Future Directions:** Our study supports the integration of automated CT-based body composition analysis into routine preoperative workflows. As most colectomy patients already undergo abdominal imaging pre-operatively, this approach introduces no additional appointment and time burden or increased utilization of radiology services. Moving forward, we must determine the best ways to scale-up and implement imaging-based body composition analysis nationally and globally. Future work must focus on determining best-practices for using body composition analysis to evaluate perioperative risk, streamlining the process, and improving user experience. In summary, this study demonstrates that automated body composition features, specifically muscle density (SMD) at L3 level, can improve surgical risk prediction in patients undergoing colectomy. These metrics contribute independent predictive power when combined with standard clinical variables or even with the existing NSQIP surgical risk calculator and are particularly useful for risk stratifying patients identified by surgeons as low-risk and identifying higher-than-expected risk patients overlooked by existing risk predictors.

**Limitations:** This study has some limitations. First, despite including data from three sites, the validation set was geographically limited and relatively small. To mitigate this, image-based scores were further proved effective by incorporating into a well-established risk model instead of building a complicated multivariable model from scratch. Second, while integrating body composition scores into risk prediction is promising, only 48% of patients had preoperative abdominal CT within 90 days.

## Conclusion

This study demonstrates that automated imaging-based frailty metrics can enhance preoperative risk stratification in colectomy patients, particularly in refining risk assessment for patients overlooked by traditional clinical models. Incorporating selected imaging biomarkers could provide incremental value, especially in clinically low-risk patients. Future work should explore integration into clinical decision-support tools and the effect their use has on pre-operative decision making, planned post-operative care, and patient outcomes.

## Acknowledgement

No conflict of interest

# Tables

Table 1 Results of univariate predictions and feature selection results for image-based frailty scores in assessing *1-year mortality, 30-day mortality, any complication,* and *serious complication,* details of the procedure are shown in [Figure 1, part(b)] left two boxes. The "Vertebral Level" column indicates the optimal vertebral level for measuring this score to predict each outcome. The "P-adjusted" value reflects the statistical significance of the variable after adjusting for BMI category, age category, functional status, smoking status, and ASA class. A dash (-) signifies that the score is either not predictive of the outcome or was automatically removed due to high correlation with confounders. If marked "Yes," the score was selected for multivariable modeling of that outcome. Notably, for mortality predictions, a single survival model was used for both 1-year and 30-day mortality, with the final vertebral level selection based on the optimal level identified for *1-year mortality*.

| | *1-year Mortality* | | | | | | *30-day mortality* | | | | | | Final selection |
|---|---|---|---|---|---|---|---|---|---|---|---|---|---|
| Image-based metric | Vertebral level | Odds ratio [5%, 95% CI] | AUC | p-value | Adjusted p-value | If predictive &independent | Image-based metric | Vertebral level | Odds ratio [5%, 95% CI] | AUC | p-value | Adjusted p-value | If predictive &independent | |
| **N_SMD** | L3 | 0.42 [0.30, 0.58] | 0.71 | 0.000 | 0.000 | **yes** | N_SMD | 3D | 0.33 [0.20,0.54] | 0.78 | 0.000 | 0.002 | **yes** | N_SMD (L3) |
| **N_SMA** | L3-L4 | 0.52 [0.33, 0.82] | 0.63 | 0.005 | 0.215 | - | N_SMI | L3 | 0.80 [0.43,1.45] | 0.63 | 0.46 | 0.4 | - | - |
| **N_MFA** | L1 | 0.72 [0.507,1.024] | 0.60 | 0.07 | 0.155 | - | N_MFA/SMA | L3 | 1.64 [1.12,2.4-] | 0.61 | 0.01 | 0.04 | **yes** | - |
| **N_VFA/SFA** | L1 | 0.72 [0.5, 1.04] | 0.60 | 0.08 | 0.012 | **yes** | N_VFA/SFA | L1 | 0.54 [0.29,1.00] | 0.64 | 0.05 | 0.05 | **yes** | N_FVA/SFA(L1) |

| Image-based metric | Vertebral level | Odds ratio [5%, 95% CI] | AUC | p-value | p-adjusted | If predictive &independent | Image-based metric | Vertebral level | Odds ratio [5%, 95% CI] | AUC | p-value | p-adjusted | If predictive &independent |
|---|---|---|---|---|---|---|---|---|---|---|---|---|---|
| **N_VFA/BODY** | L1_L2 | 0.72 [0.5,1.04] | 0.59 | 0.09 | 0.05 | yes | **N_VFA/BODY** | L1_L2 | 0.67 [0.38,1.18] | 0.60 | 0.17 | 0.158 | - |
| **SFA/BODY** | L4 | 0.03 [0.002,9.859] | 0.585 | 0.04 | 0.5 | - | **SFA/BODY** | L4 | 0.03 [0.00,2.89] | 0.59 | 0.13 | 0.215 | - |
| **MFR** | L1_L2 | 1.21[0.85,1.74] | 0.57 | 0.29 | 0.05 | - | **MFR** | L1_L2 | 1.35 [0.86,2.13] | 0.55 | 0.19 | 0.01 | - |
| **N_SFA** | L4 | 0.81 [0.58, 1.15] | 0.56 | 0.24 | 0.661 | - | **N_SFA** | 3D | 1.15 [0.75,1.75] | 0.53 | 0.51 | 0.446 | - |

| *Any complication* | | | | | | | *Serious complication* | | | | | | |
|---|---|---|---|---|---|---|---|---|---|---|---|---|---|
| Image-based metric | Vertebral level | Odds ratio [5%, 95% CI] | AUC | p-value | p-adjusted | If predictive &independent | Image-based metric | Vertebral level | Odds ratio [5%, 95% CI] | AUC | p-value | p-adjusted | If predictive &independent |
| **N_SMFD** | T12 | 0.62 [0.49, 0.78] | 0.63 | 0.000 | 0.000 | yes | **N_SMFD** | T12 | 0.66 [0.53, 0.83] | 0.61 | 0.000 | 0.000 | yes |
| **VFA** | L4 | 1.00 [1.00, 1.00] | 0.63 | 0.000 | 0.006 | yes | **N_VFA** | L4 | 1.41 [1.14, 1.74] | 0.61 | 0.001 | 0.000 | yes |
| **N_MFA/SFA** | T12-L1 | 1.96 [1.00, 3.86] | 0.58 | 0.05 | 0.82 | - | **MFA** | T12-L1 | 1.33 [1.08,1.62] | 0.58 | 0.06 | 0.29 | - |
| **N_MFA** | T12 | 1.21 [0.99,1.50] | 0.58 | 0.06 | 0.04 | yes | **N_MFR** | L4 | 0.74 [0.53, 1.04] | 0.60 | 0.08 | 0.02 | yes |
| **N_SFA** | T12 | 1.28 [1.05, 1.56] | 0.57 | 0.02 | 0.03 | yes | **N_SFA** | T12 | 1.33 [1.08, 1.62] | 0.58 | 0.005 | 0.005 | yes |
| **N_SOI** | L3 | 0.58 [0.34, 0.98] | 0.57 | 0.04 | 0.03 | yes | **-** | - | - | - | - | - | - |
| **VFA/SFA** | L4 | 1.49 [0.98, 2.27] | 0.57 | 0.06 | 0.29 | - | **VFA/SFA** | L4 | 1.38 [0.97, 2.27] | 0.56 | 0.07 | 0.21 | - |
| **SFA/BODY** | T12 | 10.94 [1.17, 102.52] | 0.55 | 0.04 | 0.60 | - | **SFA/BODY** | T12 | 16.23 [1.68, 156.39] | 0.56 | 0.02 | 0.74 | - |

**Figures**

Figure 1 Pipeline of the algorithm development and validation, including (a) image score extraction algorithm1, including both 2D scores and 3D scores extraction branch; (b) development of image-based predictors using the develop cohort; (c) validating of these predictors in an independent validate cohort.

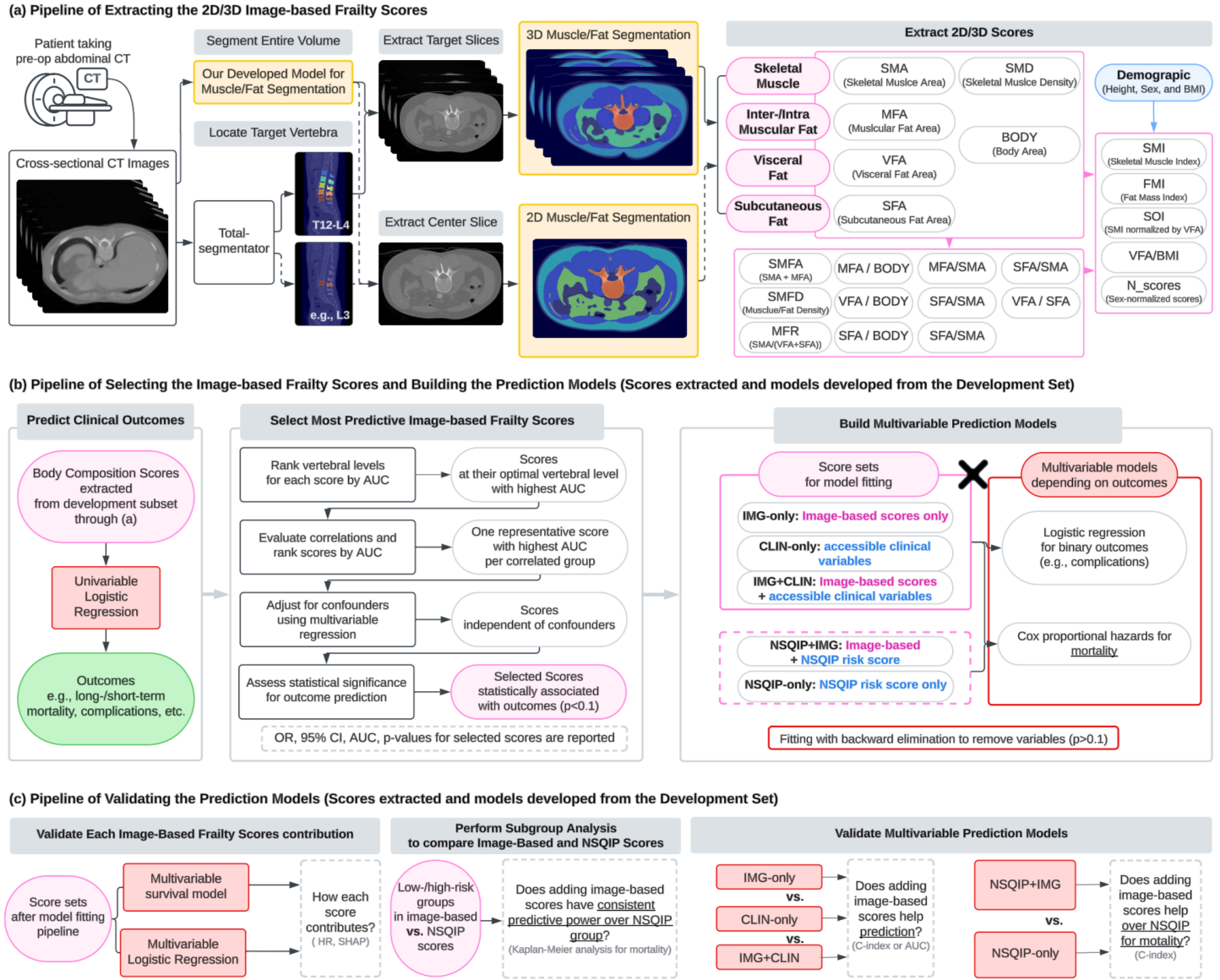

Figure 2 (a) AUC of univariate regression models based on automatically extracted image-based scores to predict 1-year mortality, 30-day mortality on the development set, respectively; (b) 30-day any complication, serious complications, and 4 different types of main complications on the development set, respectively.

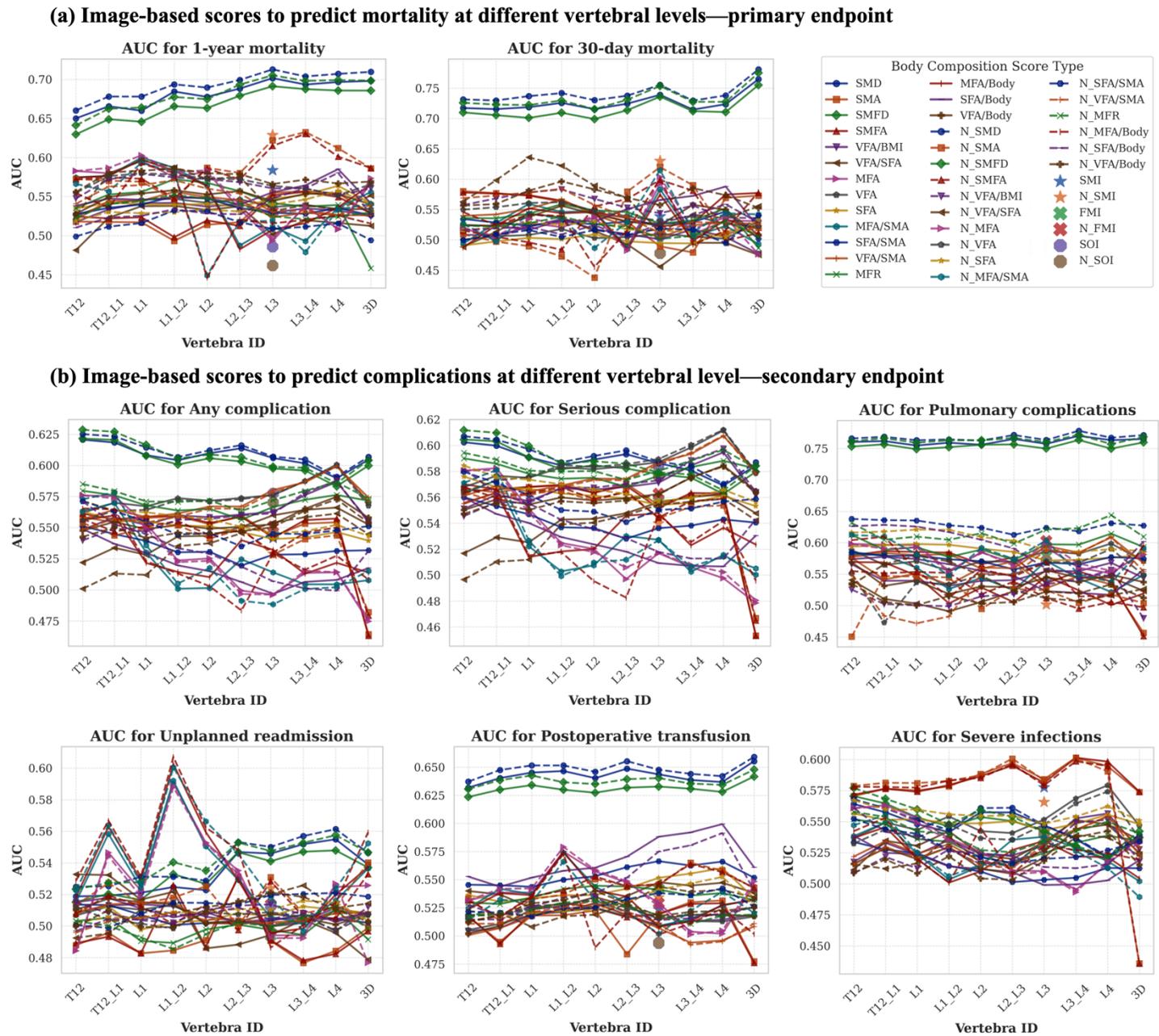

Figure 3: Model interpretability and the validation performance across outcomes. The top three rows show the contribution of each input variable to the model's prediction, visualized separately for: (1) imaging-only (IMG-only), (2) clinical-only (CLIN-only), and (3) combined imaging and clinical input (IMG+CLIN) models.

These include hazard ratios (HR) for mortality and SHAP value plots for other outcomes. The bottom row displays validation set performance metrics: C-index and integrated Brier score (IBS) for the Cox model (mortality), and AUC and Brier score for logistic regression models (complication outcomes). Error bars represent 95% confidence intervals. Severe infection and Unplanned readmission were excluded in this validation due to a lack of significant image-based predictors-none of the image-based scores left after feature selection. Numerical numbers for each metrics are listed in eTable 4.

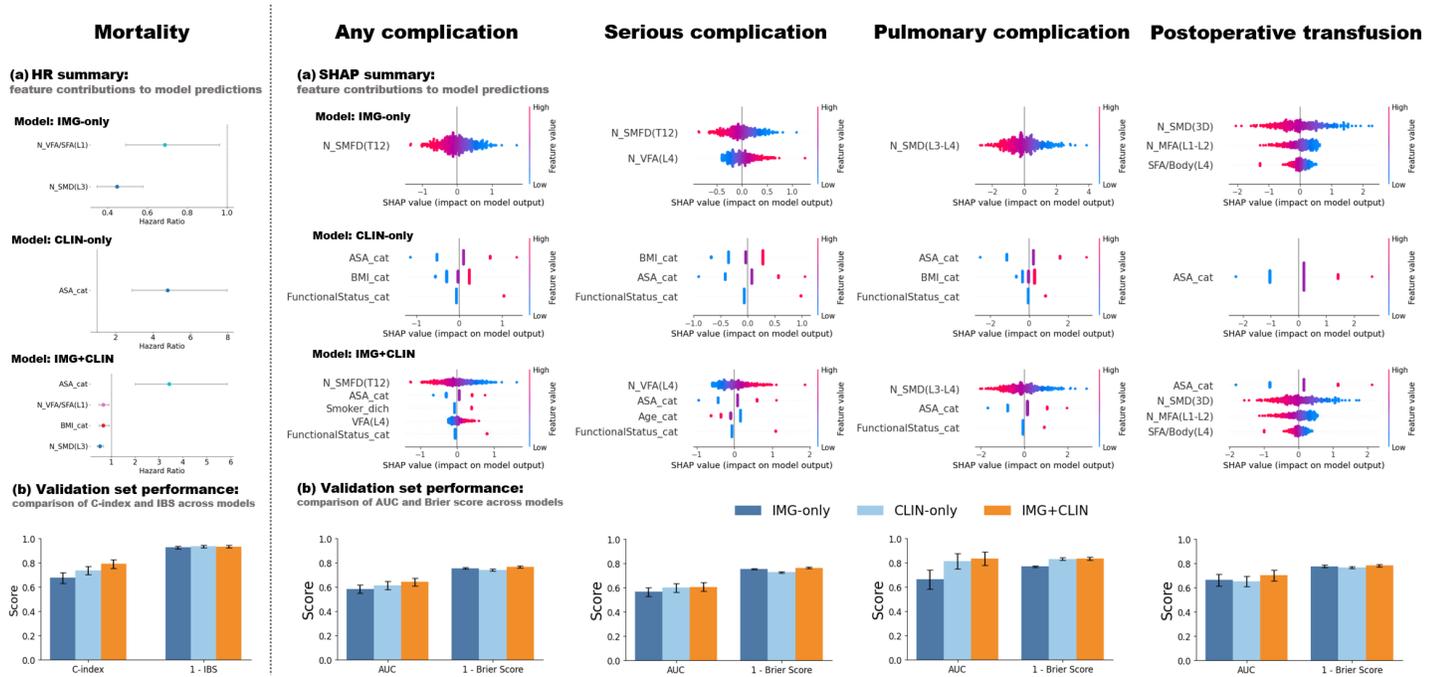

Figure 4 Feature Contributions and Risk Stratification Performance compared with NSQIP risk predictor. (a) Hazard ratio summary showing the relative contribution of NSQIP-mortality and muscle density (N_SMD(L3)) to overall mortality using Cox proportional hazards models for two multivariable models (IMG+NSQIP and NSQIP-only). (b) C-statistic curves showing the discriminative performance of IMG+NSQIP vs NSQIP-only models across varying NSQIP-mortality thresholds. (c) Kaplan-Meier survival analysis stratified by combinations of high vs low NSQIP risk and muscle density. Log-rank p-values indicate significant differences in postoperative survival across strata.

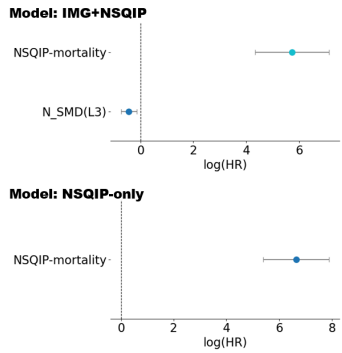
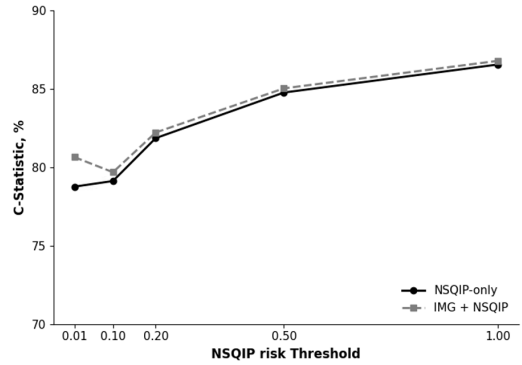
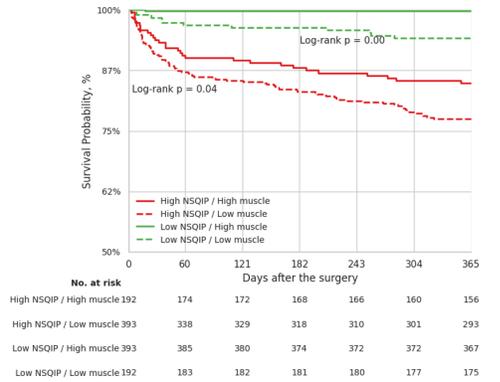

*Supplementary materials*

**eFigure 1 Data inclusive/exclusive criteria.**

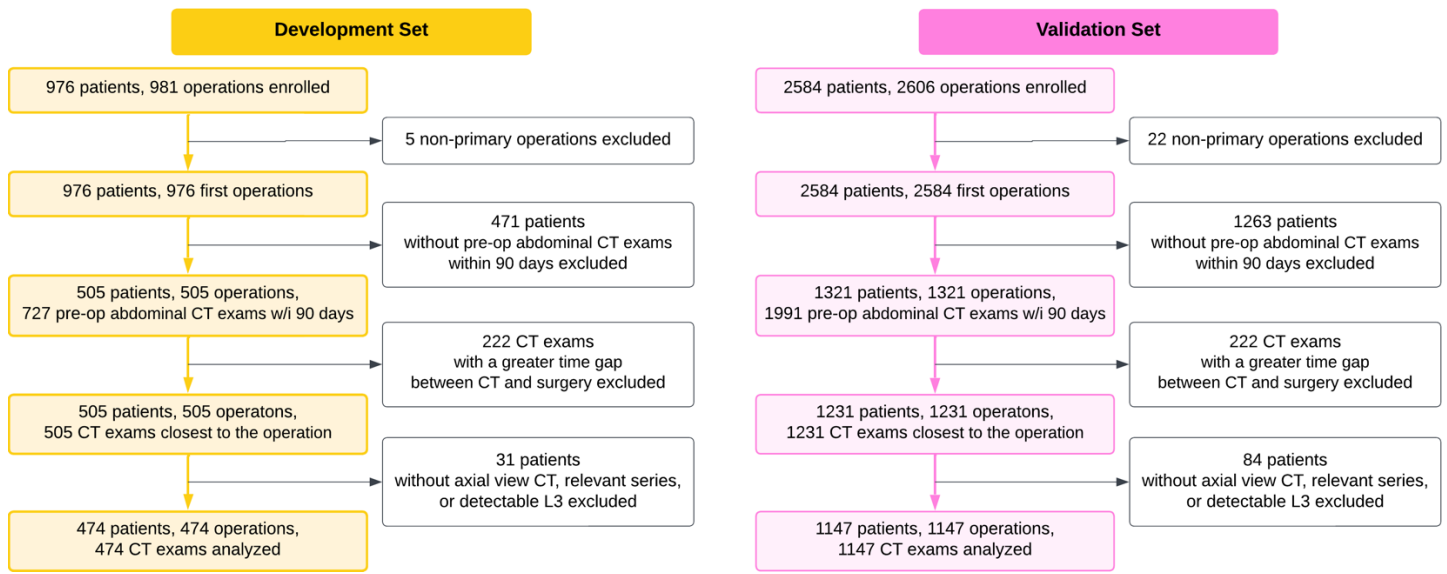

**eMethod 1 Automatic body composition segmentation.**

We performed automated segmentation of body composition using a previously validated CT segmentation model developed by Chen et al., This deep-learning-based model is capable of segmenting skeletal muscle, visceral fat, subcutaneous fat, and intra-/inter-muscular fat at the pixel level in cross-sectional CT exams. This model was built upon 2D CT slices covering the region from the upper chest to the hip and demonstrated the state-of-art performance across both internal and external datasets. To identify anatomical planes for score extraction, we also utilized TotalSegmentor[1], a publicly available deep-learning based multi-structure segmentation tool, to segment and locate vertebrae.

**eTable 1 List of all body composition scores evaluated in this study.**

| Score Name | Description | Calculation |
|---|---|---|
| **Direct Scores** | | |
| SMA | skeletal muscle area | segmented skeletal muscle area (mm^2) measured at T12, L1, …, L4 |
| SMD | skeletal muscle density | mean HU of segmented skeletal muscle measured at T12, L1, …, L4 |
| SFA | subcutaneous fat area | segmented subcutaneous fat area (mm^2) measured at T12, L1, …, L4 |
| VFA | visceral fat area | segmented visceral fat area (mm^2) measured at T12, L1, …, L4 |
| MFA | inter-/intra-muscular fat area | segmented inter-/intra-muscular fat area (mm^2) measured at T12, L1, …, L4 |
| BODY | body area | all non-background pixels with HU > -1000 at T12, L1, …, L4 |
| SMA_3D | skeletal muscle volume in 3D | total segmented skeletal muscle volume (mm^3) from T12 to L4 |
| SMD_3D | skeletal muscle density in 3D | mean HU of segmented skeletal muscle from T12 to L4 |
| SFA_3D | subcutaneous fat volume in 3D | total segmented subcutaneous fat volume (mm^3) from T12 to L4 |
| VFA_3D | visceral fat volume in 3D | total segmented visceral fat volume (mm^3) from T12 to L4 |
| MFA_3D | inter-/intra-muscular fat volume in 3D | total segmented inter-/intra-muscular fat volume (mm^3) from T12 to L4 |
| BODY_3D | body area in 3D | all non-background pixels with HU > -1000 from T12 to L4 |
| **Derived Scores: Within-Body Ratios** | | |
| SMFA | combined skeletal muscle and inter-/intra-muscular fat area | SMA + MFA at T12, L1, … L4 |
| SMFD | skeletal muscle and inter-/intra-muscular fat density | mean HU of segmented skeletal muscle and inter-/intra-muscular fat at T12, L1, … L4 |
| MFR | muscle-to-fat ratio | SMA / (VFA+SFA) at T12, L1, …, L4 |
| VFA/SFA | ratio of visceral to subcutaneous fat area | VFA / SFA at T12, L1, …, L4 |
| SFA/SMA | ratio of subcutaneous fat to skeletal muscle area | SFA / SMA at T12, L1, …, L4 |
| VFA/SMA | ratio of visceral fat to skeletal muscle area | VFA / SMA at T12, L1, …, L4 |
| MFA/SMA | ratio of inter-/intra-muscular fat to skeletal muscle area | MFA / SMA at T12, L1, …, L4 |
| SFA/BODY | ratio of subcutaneous fat to body area | SFA / BODY at T12, L1, …, L4 |
| VFA/BODY | ratio of visceral fat to body area | VFA / BODY at T12, L1, …, L4 |
| MFA/BODY | ratio of inter-/intra-muscular fat to body area | MFA / BODY at T12, L1, …, L4 |
| SMFA_3D | combined skeletal muscle and inter-/intra-muscular fat volume in 3D | SMA_3D + MFA_3D from T12 to L4 |
| SMFD_3D | mean density of skeletal muscle and inter-/intra-muscular fat in 3D | mean HU of segmented skeletal muscle and inter-/intra-muscular fat from T12 to L4 |
| MFR_3D | muscle-to-fat ratio in 3D | SMA_3D / (VFA_3D+SFA_3D) from T12 to L4 |
| VFA/SFA_3D | visceral-to-subcutaneous fat ratio in 3D | VFA_3D / SFA_3D from T12 to L4 |
| SFA/SMA_3D | subcutaneous fat-to-skeletal muscle ratio in 3D | SFA_3D / SMA_3D from T12 to L4 |
| VFA/SMA_3D | visceral fat-to-skeletal muscle ratio in 3D | VFA_3D / SMA_3D from T12 to L4 |
| MFA/SMA_3D | inter-/intra-muscular fat-to-skeletal muscle ratio in 3D | MFA_3D / SMA_3D from T12 to L4 |
| SFA/BODY_3D | subcutaneous fat-to-body area ratio in 3D | SFA_3D / body area from T12 to L4 |
| VFA/BODY_3D | visceral fat-to-body area ratio in 3D | VFA_3D / body area from T12 to L4 |
| MFA/BODY_3D | inter-/intra-muscular fat-to-body area ratio in 3D | MFA_3D / body area from T12 to L4 |
| **Derived Scores: Body-Demographic Ratios** | | |
| SMI | skeletal muscle index | SMA at L3 (cm^2) / (patient height in meters)^2 |
| FMI | fat mass (subcutaneous fat) index | SFA at L3 (cm^2) / (patient height in meters)^2 |
| SOI | Sarcopenic Obesity Index | SMI / VFA at L3 |
| VFA/BMI | visceral fat-to-BMI ratio | VFA / BMI at T12, L1, …, L4 |
| N_SMA | sex-normalized SMA | (score - avg(sex))/sd(sex)  [z-score normalization] |
| N_SMD | sex-normalized SMD | (score - avg(sex))/sd(sex)  [z-score normalization] |
| N_SFA | sex-normalized SFA | (score - avg(sex))/sd(sex)  [z-score normalization] |
| N_VFA | sex-normalized VFA | (score - avg(sex))/sd(sex)  [z-score normalization] |
| N_MFA | sex-normalized MFA | (score - avg(sex))/sd(sex)  [z-score normalization] |
| N_SMFA | sex-normalized SMFA | (score - avg(sex))/sd(sex)  [z-score normalization] |
| N_SMFD | sex-normalized SMFD | (score - avg(sex))/sd(sex)  [z-score normalization] |
| N_MFR | sex-normalized MFR | (score - avg(sex))/sd(sex)  [z-score normalization] |
| N_VFA/SFA | sex-normalized VFA/SFA | (score - avg(sex))/sd(sex)  [z-score normalization] |
| N_SFA/SMA | sex-normalized SFA/SMA | (score - avg(sex))/sd(sex)  [z-score normalization] |
| N_VFA/SMA | sex-normalized VFA/SMA | (score - avg(sex))/sd(sex)  [z-score normalization] |
| N_MFA/SMA | sex-normalized MFA/SMA | (score - avg(sex))/sd(sex)  [z-score normalization] |
| N_SFA/BODY | sex-normalized SFA/BODY | (score - avg(sex))/sd(sex)  [z-score normalization] |
| N_VFA/BODY | sex-normalized VFA/BODY | (score - avg(sex))/sd(sex)  [z-score normalization] |
| N_MFA/BODY | sex-normalized MFA/BODY | (score - avg(sex))/sd(sex)  [z-score normalization] |
| N_SMI | sex-normalized SMI | (score - avg(sex))/sd(sex)  [z-score normalization] |
| N_FMI | sex-normalized FMI | (score - avg(sex))/sd(sex)  [z-score normalization] |
| N_SOI | sex-normalized SOI | (score - avg(sex))/sd(sex)  [z-score normalization] |
| N_VFA/BMI | sex-normalized VFA/BMI | (score - avg(sex))/sd(sex)  [z-score normalization] |
| SMI_3D | skeletal muscle index in 3D | SMA_3D from T12 to L4 / (patient height in meters)^2 |
| FMI_3D | fat mass (subcutaneous fat) index in 3D | SFA_3D from T12 to L4 / (patient height in meters)^2 |
| SOI_3D | Sarcopenic Obesity Index in 3D | SMI_3D / VFA_3D from T12 to L4 |
| VFA/BMI_3D | visceral fat-to-BMI ratio in 3D | VFA_3D / BMI from T12 to L4 |
| N_SMA_3D | sex-normalized SMA in 3D | (score - avg(sex))/sd(sex)  [z-score normalization] |
| N_SMD_3D | sex-normalized SMD in 3D | (score - avg(sex))/sd(sex)  [z-score normalization] |
| N_SFA_3D | sex-normalized SFA in 3D | (score - avg(sex))/sd(sex)  [z-score normalization] |
| N_VFA_3D | sex-normalized VFA in 3D | (score - avg(sex))/sd(sex)  [z-score normalization] |
| N_MFA_3D | sex-normalized MFA in 3D | (score - avg(sex))/sd(sex)  [z-score normalization] |
| N_SMFA_3D | sex-normalized SMFA in 3D | (score - avg(sex))/sd(sex)  [z-score normalization] |
| N_SMFD_3D | sex-normalized SMFD in 3D | (score - avg(sex))/sd(sex)  [z-score normalization] |
| N_MFR_3D | sex-normalized MFR in 3D | (score - avg(sex))/sd(sex)  [z-score normalization] |
| N_VFA/SFA_3D | sex-normalized VFA/SFA in 3D | (score - avg(sex))/sd(sex)  [z-score normalization] |
| N_SFA/SMA_3D | sex-normalized SFA/SMA in 3D | (score - avg(sex))/sd(sex)  [z-score normalization] |
| N_VFA/SMA_3D | sex-normalized VFA/SMA in 3D | (score - avg(sex))/sd(sex)  [z-score normalization] |
| N_MFA/SMA_3D | sex-normalized MFA/SMA in 3D | (score - avg(sex))/sd(sex)  [z-score normalization] |
| N_SFA/BODY_3D | sex-normalized SFA/BODY in 3D | (score - avg(sex))/sd(sex)  [z-score normalization] |
| N_VFA/BODY_3D | sex-normalized VFA/BODY in 3D | (score - avg(sex))/sd(sex)  [z-score normalization] |
| N_MFA/BODY_3D | sex-normalized MFA/BODY in 3D | (score - avg(sex))/sd(sex)  [z-score normalization] |
| N_SMI_3D | sex-normalized SMI in 3D | (score - avg(sex))/sd(sex)  [z-score normalization] |
| N_FMI_3D | sex-normalized FMI in 3D | (score - avg(sex))/sd(sex)  [z-score normalization] |
| N_SOI_3D | sex-normalized SOI in 3D | (score - avg(sex))/sd(sex)  [z-score normalization] |
| N_VFA/BMI_3D | sex-normalized VFA/BMI in 3D | (score - avg(sex))/sd(sex)  [z-score normalization] |

### eMethod 2 Data collection details

For 1-year mortality, for binary category, patients were classified as follows: (1) Deceased: If a patient's status was recorded as "died" within 1-year post-operation; (2) Alive: If recorded as "alive" at their most recent follow-up, with this visit occurring beyond 1-year post-operation; (3) Unknown: If recorded as 'alive' at the most recent follow-up, but this visit occurred within 1-year post-operation, leaving their long-term survival status uncertain. For survival analysis, patient status was tracked until the last available follow-up, with mortality data censored at the end of 2023 in accordance with IRB protocol.

eTable 2 Cohort Characteristics: for some subgroups, if they are not added as N in total, it means there are some missing data for this characteristic (N/A).

| Variable | Development cohort (entire cohort) (N=976) | Development subset with CT paired (N=474) (2010-2015) | Validation cohort (entire cohort) (N=2584) | Validation subset with CT paired (N=1147) (2016-2023) |
|---|---|---|---|---|
| **Demographics** | | | | |
| - Age (years, mean+-SD) | 61.35+-14.28 | 62.15+-14.21 | 61.61+-14.55 | 61.70+-14.92 |
| o Age<65, n (%) | 530 (54.30) | 251 (52.95) | 1393 (53.91) | 617 (53.79) |
| o 65<=Age<=75 | 280 (28.69) | 138 (29.11) | 744 (28.79) | 319 (27.81) |
| o 75<Age<=85 | 139 (14.24) | 71 (14.98) | 368 (14.24) | 173 (15.08) |
| o Age>85 | 27 (2.77) | 14 (2.95) | 79 (3.06) | 38 (3.31) |
| - Male Sex, n (%) | 469 (48.05) | 231 (48.73) | 1206 (46.67) | 526 (45.86) |
| - Race, n (%) | | | | |
| o White | 712 (72.95) | 331 (69.83) | 1789 (69.23) | 760 (66.26) |
| o Black | 233 (23.87) | 125 (26.37) | 628 (24.30) | 308 (26.85) |
| o Other | 31 (3.18) | 18 (3.80) | 167 (6.46) | 79 (6.89) |
| **Characteristics/baseline risk factors** | | | | |
| - BMI (mean+-SD) | 28.19+-6.14 | 28. 39+-6.57 | 28.50+-6.80 | 28.14+-7.01 |
| o BMI<18.5, n (%) | 26 (2.66) | 17 (3.59) | 71 (2.75) | 46 (4.01) |
| o 18.5-24.99 | 294 (30.12) | 141 (29.75) | 758 (29.33) | 353 (30.78) |
| o 25-29.99 | 316 (32.38) | 143 (30.17) | 817 (31.62) | 344 (29.99) |
| o BMI>=30 | 337 (34.53) | 172 (36.29) | 907 (35.10) | 384 (33.48) |
| - Functional Status - Non-Independent, n (%) | 33 (3.38) | 20 (4.22) | 70 (2.71) | 40 (3.49) |
| - ASA Class, n (%) | | | | |
| o ASA 1 | 10 (1.03) | 3 (0.63) | 19 (0.74) | 8 (0.70) |
| o ASA 2 | 297 (30.43) | 128 (27.00) | 749 (28.99) | 298 (25.98) |
| o ASA 3 | 593 (60.76) | 294 (62.03) | 1604 (62.07) | 711 (61.99) |
| o ASA 4 | 74 (7.58) | 47 (9.92) | 199 (7.70) | 121 (10.55) |
| o ASA 5 | 1 (0.10) | 1 (0.21) | 13 (0.50) | 9 (0.78) |
| - Smoking Status (within 1 year): yes, n (%) | 162 (16.60) | 81 (17.09) | 384 (14.86) | 197 (17.18) |
| **Comorbidities, n (%)** | | | | |
| - Hypertension requiring medication: yes | 505 (51.74) | 247 (52.11) | 1295 (50.12) | 590 (51.44) |
| - Diabetes | 148 (15.16) | 83 (17.51) | 429 (16.60) | 192 (16.74) |
| - Congestive heart failure | 9 (0.92) | 8 (1.69) | 85 (3.29) | 41 (3.57) |
| - COPD | 43 (4.41) | 26 (5.49) | 137 (5.30) | 71 (6.19) |
| - Disseminated cancer | 71 (7.27) | 38 (8.02) | 154 (5.96) | 77 (6.71) |
| - Type of colectomy, n (%) | | | | |
| o Laparoscopic colorectal procedure | 643 (65.88) | 288 (60.76) | 1722 (66.64) | 622 (54.23) |
| o Open colorectal procedure | 333 (34.12) | 186 (39.24) | 863 (33.36) | 525 (45.77) |
| - Emergency surgery: yes, n (%) | 112 (11.48) | 77 (16.24) | 280 (10.84) | 211 (18.40) |
| - Manufacturer, n (%) | | | | |
| o GE | - | 301 (63.50) | - | 542 (47.25) |
| o Siemens | - | 173 (36.50) | - | 585 (51.00) |
| o Philips | - | 0 (0.00) | - | 6 (0.52) |
| o Other (Canon, etc) | - | 0 (0.00) | - | 14 (1.22) |
| - Slice thickness (mm, mean+-SD) | - | 4.24+-1.64 | - | 2.75+-1.54 |
| - Year of data collection | 2010-2015 | 2010-2015 | 2016-2023 | 2016-2023 |
| - Institutions, n | | | | |
| o DUH | 976 | 474 | 1626 | 630 |
| o DRAH | 0 | 0 | 536 | 266 |
| o DRH | 0 | 0 | 422 | 251 |

### eTable 3 Postoperative Outcomes

| Cohort Group | Long Term Outcomes (1-year) | Short Term Outcomes (30 days) | | | | | | | | | | | | | |
|---|---|---|---|---|---|---|---|---|---|---|---|---|---|---|---|
| | Mortality | Mortality | Any complication | Serious complication | Unplanned readmission | Unplanned return to OR | Pulmonary complication | Cardiac complication | Sepsis | Septic shock | Cdiff | Renal complication | Neurological events | Thromboembolic events | Post-operative transfusion | Severe infections |
| **Develop cohort – entire (N=976), (n, %)** | | | | | | | | | | | | | | | | |
| yes | 62 (6.35) | 26 (2.66) | 217 (22.13) | 208 (21.31) | 121 (12.39) | 34 (3.48) | 51 (5.22) | 16 (1.64) | 40 (4.10) | 19 (1.9) | 2 (0.2) | 18 (1.84) | 8 (0.82) | 22 (2.25) | 126 (12.91) | 115 (11.78) |
| no | 785 (80.43) | 924 (94.67) | 759 (77.8) | 768 (78.69) | 855 (87.60) | 942 (96.5) | 925 (94.77) | 960 (98.36) | 936 (95.9) | 957 (98.1) | 974 (99.8) | 958 (98.16) | 968 (99.18) | 954 (97.75) | 850 (87.09) | 861 (88.22) |
| unknown | 129 (13.22) | 26 (2.66) | - | - | - | - | - | - | - | - | - | - | - | - | - | - |
| **Develop cohort – subset (N=474), (n, %)** | | | | | | | | | | | | | | | | |
| yes | 45 (9.49) | 19 (4.01) | 119 (25.1) | 114 (24.05) | 66 (13.92) | 14 (5.1) | 31 (6.54) | 12 (2.53) | 28 (2.9) | 12 (2.5) | 1 (0.2) | 13 (2.74) | 2 (0.42) | 12 (2.53) | 78 (16.46) | 60 (12.66) |
| no | 376 (79.32) | 445 (93.88) | 355 | 360 (75.94) | 408 (86.08) | 460 | 443 (93.45) | 462 (97.47) | 446 | 462 | 473 | 461 (97.26) | 472 (99.58) | 462 (97.49) | 396 (83.54) | 414 (87.34) |
| unknown | 53 (11.18) | 10 (2.11) | - | - | - | - | - | - | - | - | - | - | - | - | - | - |
| **Validation cohort – entire (N=2584), (n, %)** | | | | | | | | | | | | | | | | |
| yes | 186 (7.20) | 72 (2.79) | 685 (25.61) | 590 (20.67) | 274 (10.60) | 128 (49.53) | 95 (3.68) | 19 (0.74) | 87 (3.4) | 81 (3.1) | 26 (1.0) | 101 (3.91) | 3 (0.12) | 46 (1.78) | 258 (9.98) | 198 (7.66) |
| no | 2235 (86.49) | 2483 (96.09) | 1899 (73.49) | 1994 (77.17) | 2310 (89.40) | 2456 (95.05) | 2489 (96.32) | 2565 (99.26) | 2497 | 2503 | 2558 | 2483 (96.09) | 2581 (99.88) | 2538 (98.2) | 2326 (90.02) | 2386 (92.34) |
| unknown | 163 (6.31) | 29 (1.12) | - | - | - | - | - | - | - | - | - | - | - | - | - | - |
| **Validation cohort – subset (N=1147), (n, %)** | | | | | | | | | | | | | | | | |
| yes | 131 (11.42) | 56 (4.88) | 381 (33.2) | 330 (28.77) | 125 (10.90) | 66 (5.8) | 62 (5.40) | 7 (0.61) | 53 (4.6) | 53 (4.6) | 18 (1.6) | 54 (4.71) | 2 (0.17) | 24 (2.09) | 143 (12.47) | 104 (9.07) |
| no | 959 (83.61) | 1074 (93.54) | 766 (66.78) | 817 (71.22) | 1022 (89.10) | 1081 | 1085 (94.59) | 1140 (99.40) | 1094 | 1094 | 1129 | 1093 (95.29) | 1145 (98.83) | 1123 (97.91) | 1004 (87.53) | 1043 (90.93) |
| unknown | 57 (4.97) | 17 (1.48) | n/a | n/a | n/a | n/a | n/a | n/a | n/a | n/a | n/a | n/a | n/a | n/a | n/a | n/a |

eTable 4 Independent analysis for image-based frailty scores and cofounders, as well as the adjusted p-value to predict mortality and morbidity.

| Variable name | Vertebral level | Cofounders using ANOVA test p<0.01 (YES YES ), p<0.1 (YES ) | | | | |
|---|---|---|---|---|---|---|
| | | Functional Status | BMI_cat | Age_cat | Smoker | ASA_cat |
| **N_SMD** | L3 | YES | YES YES | YES YES | YES YES | YES YES |
| **N_SMA** | L3_L4 | | YES YES | YES YES | | |
| **N_MFA** | L1 | | YES YES | | | YES |
| **N_VFA/SFA** | L1 | | YES | YES YES | YES YES | |
| **N_VFA/BODY** | L1_L2 | | YES YES | YES YES | YES YES | YES |
| **SFA/BODY** | L4 | | YES YES | YES YES | YES YES | YES YES |
| **MFR** | L1_L2 | | YES YES | YES | YES YES | YES YES |
| **N_SFA** | L4 | | YES YES | YES YES | YES | |

**eMethod 3 Comparison with NSQIP risk scores on the development set**

For those image-based frailty scores selected for mortality model, N_SMD (L3) exhibited a moderate correlation with NSQIP mortality risk prediction (Pearson: -0.35; Spearman: -0.46, p < 0.001), and SFA/BODY (L4) had a weaker correlation with NSQIP mortality risk (Pearson: -0.10; Spearman: -0.24, p < 0.01). For those scores selected for any complication model, N_SMFD (T12) showed a moderate correlation with NSQIP any complication risk prediction (Pearson: -0.28; Spearman: -0.27, p < 0.001), VFA (L4) had a weaker correlation

(Pearson: -0.08; Spearman: -0.14, p < 0.1), and N_MFA (T12) (Pearson: 0.041, p=0.4; Spearman: 0.104, p < 0.1), N_SFA (Pearson: 0.08, p=0.1; Spearman: 0.12, p < 0.1), N_SOI (Pearson: -0.05, p=0.3; Spearman: -0.1, p < 0.1), had barely or almost no correlation to NSQIP any complication risk prediction.

eTable 4 Input1 sets and validation performance for the predictive models: Input variables for single-variable and multivariable predictive models of mortality and postoperative complications. The single-score model includes only the most predictive image-based frailty scores after confounder adjustment. Multivariable models incorporate image-based scores, clinical confounders, or both, with variable selection performed via automatic backward elimination (variables with p>0.1 were removed). *Severe infection* and *Unplanned readmission* were excluded in this validation due to a lack of significant image-based predictors-none of the image-based scores left after feature selection. P-value was reported based on the bootstrap resampling test.

| Model name | Model version | *Mortality* | | | *Any complication* | | |
|---|---|---|---|---|---|---|---|
| | | Inputs | C-index | Brier score | inputs | AUC | Brier score |
| IMG-only | Multivariable | N_SMD (L3) | 0.70 [0.66,0.75) | 0.07 [0.06,0.08] | N_SMFD(T12) | 0.58 [0.55,0.62] | 0.24 |
| CLIN-only | Multivariable | ASA class | 0.75 [0.72,0.79] | 0.05 [0.05,0.07] | Functional status, ASA_cat, BMI cat | 0.61 [0.58,0.65] | 0.26 |
| IMG+CLIN | Multivariable | N_SMD (L3), BMI cat, Age cat | **0.80 [0.77,0.84] (p<0.001)** | 0.06 [0.05,0.07] | N_SMFD (T12), VFA(L4), Functional status, Smoker, ASA cat | **0.64 [0.61,0.68] (p=0.06)** | 0.23 |
| Model name | Model version | *Serious complication* | | | *Pulmonary complication* | | | *Postoperative transfusion* | | |
| | | inputs | AUC | Brier Score | inputs | AUC | Brier Score | inputs | AUC | Brier Score |
| IMG-only | Multivariable | N_SMFD (T12), N_VFA (L4) | 0.57 [0.53,0.60] | 0.25 | N_SMD (L3-L4) | 0.66 [0.60-0.73] | 0.20 | N_SMD (3D), SFA/Body (L4), N_MFA (L1-L2) | 0.66 [0.62,0.71] | **0.221 [0.21,0.23]** |
| CLIN-only | Multivariable | Functional status, ASA cat BMI cat | 0.6 [0.56,0.64] | 0.27 | ASA class, BMI cat, Functional status | 0.76 [0.69,0.83] | 0.25 | ASA cat | 0.65 [0.61,0.69] | **0.235 [0.226,0.243]** |

| IMG+CLIN | Multivariable | N_VFA (L4), Functional status, ASA class, Age cat | **0.61 [0.57,0.64]** | 0.23 | N_SMD (L3-L4), ASA class, functional status | **0.78 [0.71,0.84]** | 0.18 | N_SMD (3D), SFA/Body (L4), N_MFA (L1-L2), ASA_cat | **0.70 [0.66,0.75] (p<0.001)** | 0.217 |